\documentclass{sig-alternate-05-2015}

\usepackage{tikz}
\usetikzlibrary{automata}

\usepackage{bbm}
\usepackage{amsmath,amsfonts,amssymb}
\usepackage{subfigure,enumitem}

\usepackage{url}
\usepackage[naturalnames,colorlinks,
			pdfpagemode={UseOutlines},
			bookmarks, 			
			bookmarksopen=true,
			pdfstartview={FitH},
			bookmarksnumbered=true,            		
			linkcolor=blue,
            citecolor=blue,
            urlcolor=magenta,
            linktocpage,
            plainpages=false,
    		pdfauthor={Siddharth Reddy, Igor Labutov, Siddhartha Banerjee, Thorsten Joachims},
			pdftitle={Unbounded Human Learning: Optimal Scheduling for Spaced Repetition},
            pdftex
]{hyperref}

\begin{document}






\CopyrightYear{2016} 
\setcopyright{acmlicensed}
\conferenceinfo{KDD '16,}{August 13 - 17, 2016, San Francisco, CA, USA}
\isbn{978-1-4503-4232-2/16/08}\acmPrice{\$15.00}
\doi{http://dx.doi.org/10.1145/2939672.2939850}

%

\title{Unbounded Human Learning: Optimal Scheduling for Spaced Repetition}
%
%
%
%
%

\numberofauthors{4} 
%
\author{
%
%
\alignauthor
Siddharth Reddy\\
       \affaddr{Department of Computer Science}\\
       \affaddr{Cornell University}\\
       \email{sgr45@cornell.edu}
\alignauthor
Igor Labutov\\
       \affaddr{Electrical and Computer Engineering}\\
       \affaddr{Cornell University}\\
       \email{iil4@cornell.edu}
\alignauthor Siddhartha Banerjee\\
       \affaddr{Operations Research and Information Engineering}\\
       \affaddr{Cornell University}\\
       \email{sbanerjee@cornell.edu}
\and  
\alignauthor Thorsten Joachims\\
       \affaddr{Department of Computer Science}\\
       \affaddr{Cornell University}\\
       \email{tj@cs.cornell.edu}
}

\maketitle
\begin{abstract}
In the study of human learning, there is broad evidence that our ability to retain information improves with repeated exposure and decays with delay since last exposure. This plays a crucial role in the design of educational software, leading to a trade-off between teaching new material and reviewing what has already been taught. A common way to balance this trade-off is \emph{spaced repetition}, which uses periodic review of content to improve long-term retention. Though spaced repetition is widely used in practice, e.g., in electronic flashcard software, there is little formal understanding of the design of these systems. Our paper addresses this gap in three ways. First, we mine log data from spaced repetition software to establish the functional dependence of retention on reinforcement and delay. Second, we use this memory model to develop a stochastic model for spaced repetition systems. We propose a queueing network model of the \emph{Leitner system} for reviewing flashcards, along with a heuristic approximation that admits a tractable optimization problem for review scheduling. Finally, we empirically evaluate our queueing model through a Mechanical Turk experiment, verifying a key qualitative prediction of our model: the existence of a sharp phase transition in learning outcomes upon increasing the rate of new item introductions.
\end{abstract}

%
%




\section{Introduction}
\label{sec:intro}

The ability to learn and retain a large number of new pieces of information is an essential component of human learning. Scientific theories of human memory, going all the way back to 1885 and the pioneering work of Ebbinghaus \cite{ebbinghaus1913memory}, identify two critical variables that determine the probability of recalling an item: \emph{reinforcement}, i.e., repeated exposure to the item, and \emph{delay}, i.e., time since the item was last reviewed. Accordingly, scientists have long been proponents of the \emph{spacing effect} for learning: the phenomenon in which periodic, spaced review of content improves long-term retention.

A significant development in recent years has been a growing body of work that attempts to `engineer' the process of human learning, creating tools that enhance the learning process by building on the scientific understanding of human memory. These educational devices usually take the form of `flashcards' -- small pieces of information content which are repeatedly presented to the learner on a schedule determined by a \emph{spaced repetition} algorithm \cite{gwern-blog-post}. Though flashcards have existed for a while in physical form, a new generation of spaced repetition software such as SuperMemo \cite{wozniak1994optimization}, Anki \cite{anki}, Mnemosyne \cite{mnemosyne}, Pimsleur \cite{pimsleur1967memory}, and Duolingo \cite{duolingo} allow a much greater degree of control and monitoring of the review process. These software applications are growing in popularity \cite{gwern-blog-post}, but there is a lack of formal mathematical models for reasoning about and optimizing such systems. In this work, we combine memory models from psychology with ideas from queueing theory to develop such a mathematical model for these systems. In particular, we focus on one of the simplest and oldest spaced repetition methods: the \emph{Leitner system} \cite{leitner1974so}.

The Leitner system, first introduced in 1970, is a heuristic for prioritizing items for review. It is based on a series of decks of flashcards. After the user sees a new item for the first time, it enters the system at deck 1. The items at each deck form a first-in-first-out (FIFO) queue, and when the user requests an item to review, the system chooses a deck $i$ according to some schedule, and presents the top item. If the user does not recall the item, the item is added to the bottom of deck $i-1$; else, it is added to the bottom of deck $i+1$. The aim of the scheduler is to ensure that items from lower decks are reviewed more often than those from higher decks, so the user spends more time working on forgotten items and less time on recalled items. Existing schemes for assigning review frequencies to different decks are based on heuristics that are not founded on any formal reasoning, and hence, have no optimality guarantees. One of our main contributions is a principled method for determining appropriate deck review frequencies.

The problem of deciding how frequently to review different decks in the Leitner system is a specific instance of the more general problem of \emph{review scheduling} for spaced repetition software. The main challenge in all settings is that schedules must balance competing priorities of introducing new items and reviewing old items in order to maximize the rate of learning. While most existing systems use heuristics to make this trade-off, our work presents a principled understanding of the tension between novelty and reinforcement.

\subsection{Related Work}
\label{ssec:relwork}

The scientific literature on modeling human memory is highly active and dates back more than a century.
One of the simplest memory models, the \emph{exponential forgetting curve}, was first studied by Ebbinghaus in 1885 \cite{ebbinghaus1913memory} -- it models the probability of recalling an item as an exponentially-decaying function of the time elapsed since previous review and the memory `strength'. 
The exact nature of how strength evolves as a function of the number of reviews, length of review intervals, and other factors is a topic of debate, though there is some consensus on the existence of a \emph{spacing effect}, in which spaced reviews lead to greater strength than massed reviews (i.e., cramming) \cite{dempster1989spacing, cepeda2006distributed}. 
Recent studies have proposed more sophisticated probabilistic models of learning and forgetting \cite{pashler2009predicting, lindsey2014improving}, and there is a large body of related work on item response theory and knowledge tracing \cite{linden1997handbook, corbett1994knowledge}. 
Our work both contributes to this literature (via observational studies on log data from the Mnemosyne software) and uses it as the basis for our queueing model and scheduling algorithm.

Though used extensively in practice (see \cite{gwern-blog-post} for an excellent overview), there is very limited literature on the design of spaced repetition software.
One notable work in this regard is that of Novikoff et al. \cite{novikoff2012education}, who propose a theoretical framework for spaced repetition based on a set of deterministic operations on an infinite string of content pieces.
They assume identical items and design schedules to implement deterministic spacing constraints, which are based on an intuitive understanding of the effect of memory models on different learning objectives. 
The focus in \cite{novikoff2012education} is on characterizing the combinatorial properties (e.g., maximum asymptotic throughput) of schedules that implement various spacing constraints.
Though our work shares the same spirit of formalizing the spaced repetition problem, we improve upon their work in three ways: (1) in terms of empirical verification, as our work leverages both software log data and large-scale experimentation to verify the memory models we use, and test the predictions made by our mathematical models; (2) in computational terms, wherein, by using appropriate stochastic models and approximations, we formulate optimization problems that are much easier to solve; and (3) in terms of flexibility, as our model can more easily incorporate various parameters such as the user's review frequency, non-identical item difficulties, and different memory models.

\subsection{Our Contributions}
\label{ssec:contributions}

The key contributions of this paper fall into two categories. First, the paper introduces a principled methodology for designing review scheduling systems with various learning objectives. Second, the models we develop provide qualitative insights and general principles for spaced repetition. The overall argument in this paper consists of the following three steps:

\begin{enumerate}[leftmargin=*]
\item \emph{Mining large-scale log data to validate human memory models}:
First, we perform observational studies on data from Mnemosyne \cite{mnemosyne}, a popular flashcard software tool, to compare different models of retention probability as a function of reinforcement and delay.
Our results, presented in Section \ref{sec:memmodel}, add to the existing literature on memory models and provide the empirical foundation upon which we base our mathematical model of spaced repetition.
\item \emph{Mathematical modeling of spaced repetition systems}:
Our main contribution lies in embedding the above memory model into a stochastic model for spaced repetition systems, and using this model to optimize the review schedule.
Our framework, which we refer to as the \emph{Leitner Queue Network}, is based on ideas from queueing theory and job scheduling. 
Though conceptually simple and easy to simulate, the Leitner Queue Network does not provide a tractable way to optimize the review schedule. 
To this end, we propose a (heuristic) approximate model, which in simulations is close to our original model for low arrival rates, and which leverages the theory of product-form networks \cite{kelly2011reversibility,chao1999queueing} to greatly simplify the scheduling problem.
This allows us to study several relevant questions: the maximum rate of learning, the effect of item difficulties, and the effect of a learner's review frequency on their overall rate of learning.
We present our model, theory, and simulations in Section \ref{sec:queuemodel}.
\item \emph{Verifying the mathematical model in controlled experiments}: 
Finally, we use Amazon Mechanical Turk \cite{mturk} to perform large-scale experiments to test our mathematical models. 
In particular, we verify a critical qualitative prediction of our mathematical model: the existence of a \emph{phase transition} in learning outcomes upon increasing the rate of introduction of new content beyond a maximum threshold. 
Our experimental results agree well with our model's predictions, reaffirming the utility of our framework.
\end{enumerate}

Our work provides the first mathematical model for spaced repetition systems which is empirically tested and admits a tractable optimization problem for review scheduling. It opens several directions for further research: developing better models for such systems, providing better analysis for the resulting models, and performing more empirical studies to understand these systems.
We discuss some of these open questions in detail in Section \ref{sec:conclusions}.
Our experimental platform can help serve as a testbed for future studies; to this end, we release all our data and software tools to facilitate replication and follow-up studies (see Section \ref{sec:experiments}).

\section{Testing Human Memory Models}
\label{sec:memmodel}

To design a principled spaced repetition system, we must first understand how a user's ability to recall an item is affected by various system parameters.
One well-studied model of human memory from the psychology literature is the \emph{exponential forgetting curve}, which claims that the probability of recalling an item decays exponentially with the time elapsed since it was last reviewed, at a rate which decreases with the `strength' of the item's memory trace.
In this section, we conduct an observational study on large-scale log data collected from the Mnemosyne \cite{mnemosyne} flashcard software to validate the exponential forgetting curve.

\paragraph{Exponential Forgetting Curve}
\label{memory-model}

We adopt a variant of the standard exponential forgetting curve model, where recall is binary (i.e., a user either completely recalls or forgets an item) and the probability of recalling an item has the following functional form:
\begin{eqnarray}
\mathbb{P}[\mbox{recall}] = \exp{\left(-\theta \cdot d / s \right)}, \label{eq:expcurve}
\end{eqnarray}
where $\theta \in \mathbb{R}^+$ is the item difficulty, $d \in \mathbb{R}^+$ is the time elapsed since previous review, and $s \in \mathbb{R}^+$ is the memory strength. 
Our formulation is slightly different from that of Ebbinghaus \cite{ebbinghaus1913memory}, in that we have added an explicit item difficulty parameter $\theta$, which corresponds to the assumption that there is a constant, item-specific component of the memory decay rate.

To justify the use of this memory model in our scheduling algorithm, we first explore how different forms of the exponential forgetting curve model fit empirical data. 
In particular, we explore the use of a global item difficulty $\theta$ vs. an item-specific difficulty $\theta_i$, as well as several simple models of memory strength: a constant strength $s = 1$, a strength $s = n_{ij}$ equal to the number of repetitions of item $i$ for user $j$ (where $n_{ij} \geq 1$), and a strength $s = q_{ij}$ equal to the position of item $i$ in the Leitner system for user $j$ (where $q_{ij} \geq 1$).

\paragraph{Experiments on Log Data} \label{mnemosyne-experiments}

We use large-scale log data collected from Mnemosyne \cite{mnemosyne}, a popular flashcard software tool, to validate our assumptions about the forgetting curve. After filtering out users and items with fewer than five interactions, we select a random subset of the data that contains $859,591$ interactions, $2,742$ users, and $88,892$ items. 
Each interaction is annotated with a grade (on a 0-5 scale) that was self-reported by the user. 
Users are instructed by the Mnemosyne software to use a grade of 0 or 1 to indicate that they did not recall the item, and a grade of 2-5 to indicate that they did recall the item, with higher grades implying easier recall. 
We discretize grades into binary outcomes, where $recall \triangleq grade \geq 2$ and, and observe an overall recall rate of $0.56$ in the data. 
Additionally, we scale the time intervals between reviews to days.

\begin{figure}[t]
\centering
\includegraphics[width=0.9\linewidth]{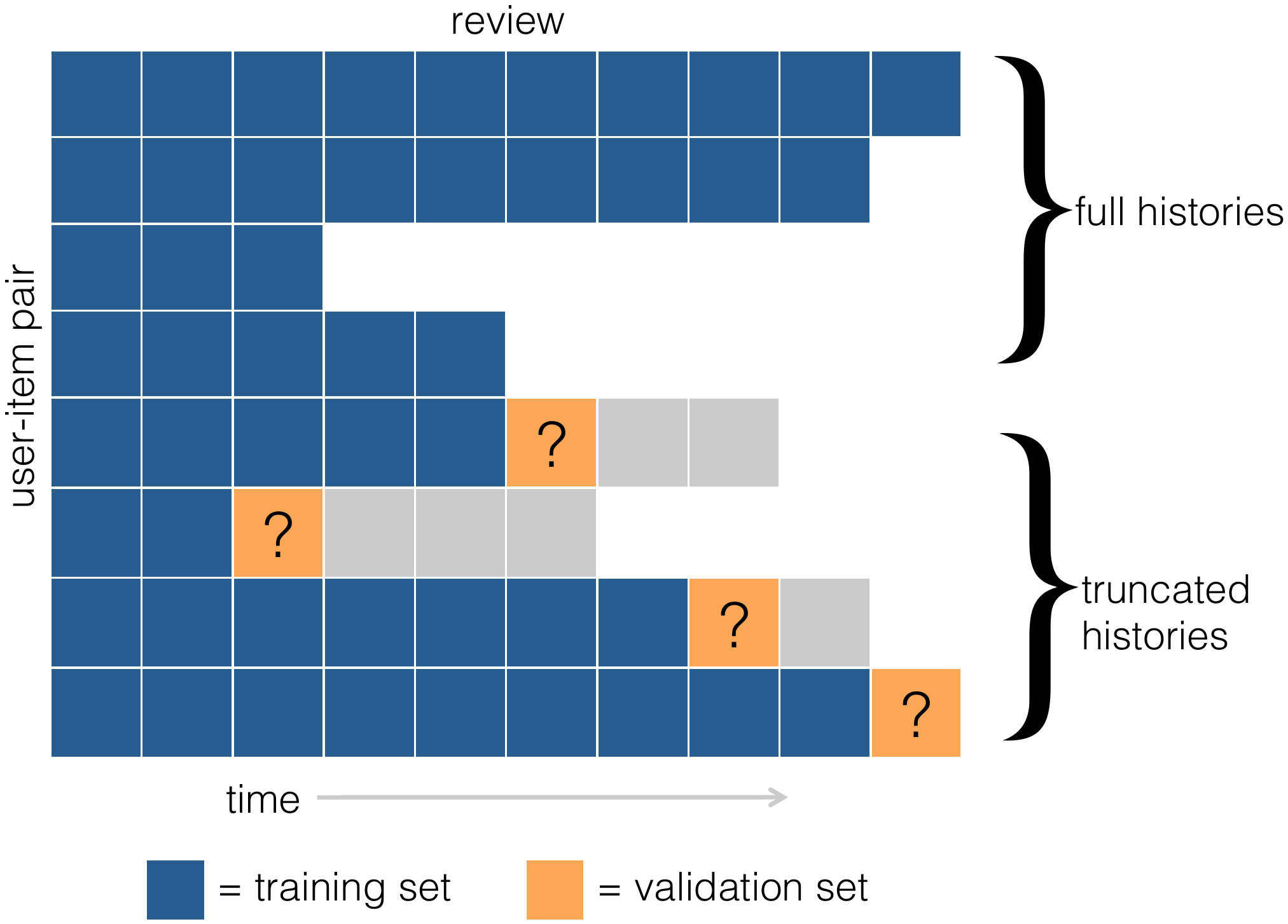}
\caption{Schematic of the classification task used to evaluate memory models: Each square corresponds to a user-item interaction with a binary outcome. The gray squares are thrown out. This training-validation split occurs on each fold, with sets of full and truncated histories changing across folds.}
\label{fig:pred-task}
\end{figure}

\begin{table}[t]
\label{mnemosyne-results}
\begin{center}
\begin{tabular}{ l l l }
    & $\mathbb{P}[\mbox{recall}]$ & Model \\ \hline
    \textbf{1} & $\phi(\theta_j)$ & 0PL-IRT \emph{user} \\
    \textbf{2} & $\phi(-\beta_i)$ & 0PL-IRT \emph{item} \\
    \textbf{3} & $\phi(\theta_j - \beta_i)$ & 1PL-IRT\\
    \textbf{4} & $\phi(\beta \cdot x)$ & Logistic regression\\ \hline
    \textbf{5} & $\exp{(-\theta \cdot d_{ij} / n_{ij})}$ & Exponential forgetting curve\\
    \textbf{6} & $\exp{(-\theta \cdot d_{ij})}$ & \\
    \textbf{7} & $\exp{(-\theta / n_{ij})}$ & \\
    \textbf{8} & $\exp{(-\theta \cdot d_{ij} / q_{ij})}$ & \\
    \textbf{9} & $\exp{(-\theta / q_{ij})}$ & \\
    \textbf{10} & $\exp{(-\theta_i \cdot d_{ij} / n_{ij})}$ & \\
    \textbf{11} & $\exp{(-\theta_i \cdot d_{ij})}$ & \\
    \textbf{12} & $\exp{(-\theta_i / n_{ij})}$ & \\
    \textbf{13} & $\exp{(-\theta_i \cdot d_{ij} / q_{ij})}$ & \\
    \textbf{14} & $\exp{(-\theta_i / q_{ij})}$ & \\
\end{tabular}
\caption{Summary of models used for prediction: In all cases, the subscripts refer to user $j$ and item $i$. Rows 1-4 represent our benchmarks; here $\phi$ is the logistic function, while in row 4, $x$ refers to the  feature vector of review interval and outcome statistics described earlier in this section, and $\beta$ is a vector of coefficients. In rows 5-14, $d_{ij}$ is the time elapsed since previous review of item $i$ for user $j$, $q_{ij}$ denotes the position of item $i$ in the Leitner system for user $j$ and $n_{ij}$ is the number of past reviews of item $i$ by user $j$. $\theta$ represents a global item difficulty, while $\theta_i$ is an item-specific difficulty for item $i$. 
}
\end{center}
\end{table}

\begin{figure}[t!]
\begin{subfigure}
\centering
\includegraphics[width=0.95\linewidth]{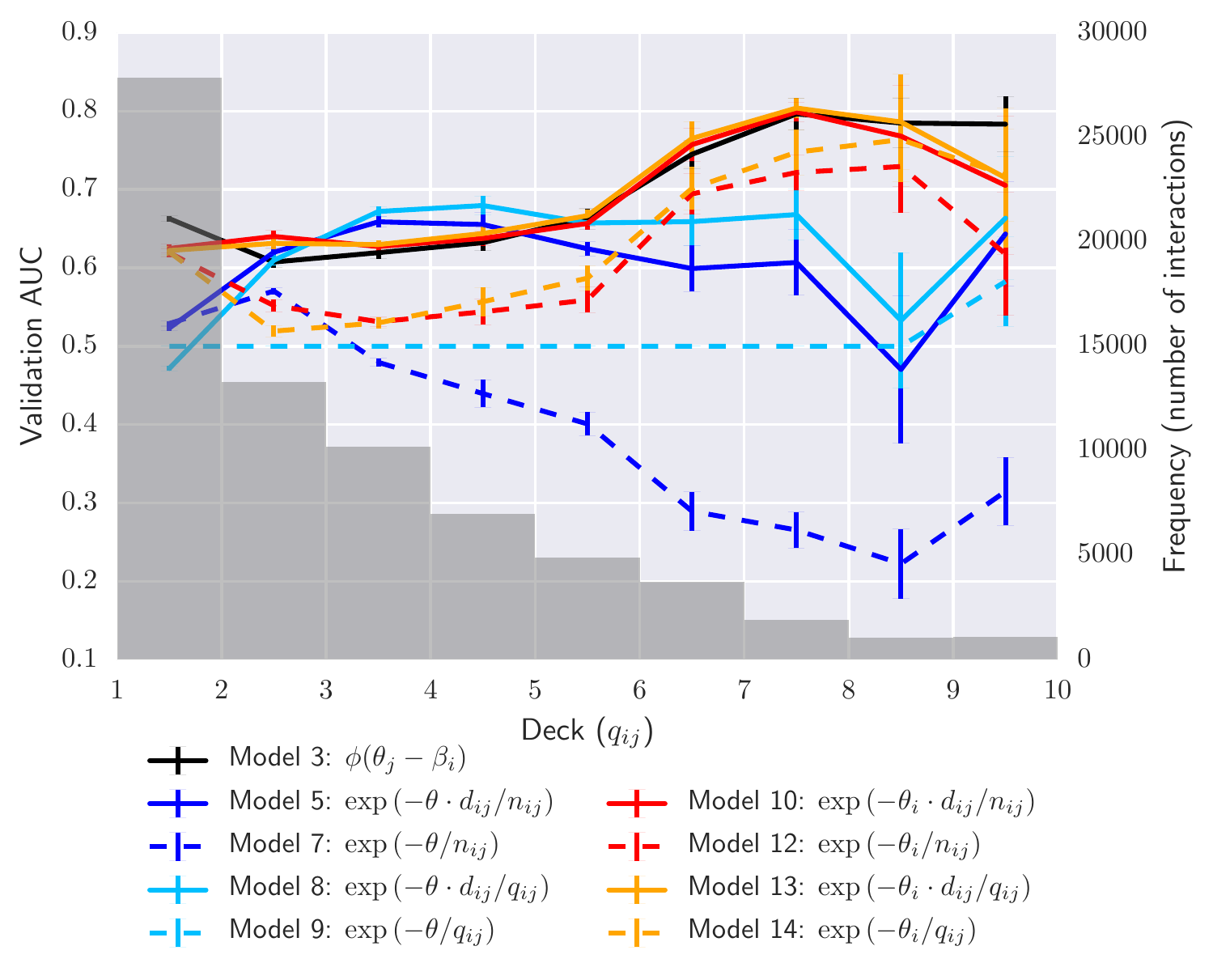}
\caption{To evaluate the memory models' ability to predict outcomes in the Leitner system, validation AUC is computed for separate bins of data that control for an item's position $q_{ij}$ in the Leitner system. The error bars show the standard error of validation AUC across the ten folds of cross-validation. Each curve corresponds to a row in Table \ref{mnemosyne-results}. We have included only the best-performing benchmark model, 1PL-IRT (model 3), to reduce clutter.}
\label{fig:auc-vs-deck}
\end{subfigure}
\vspace{0.2in}
\begin{subfigure}
\centering
\includegraphics[width=0.95\linewidth]{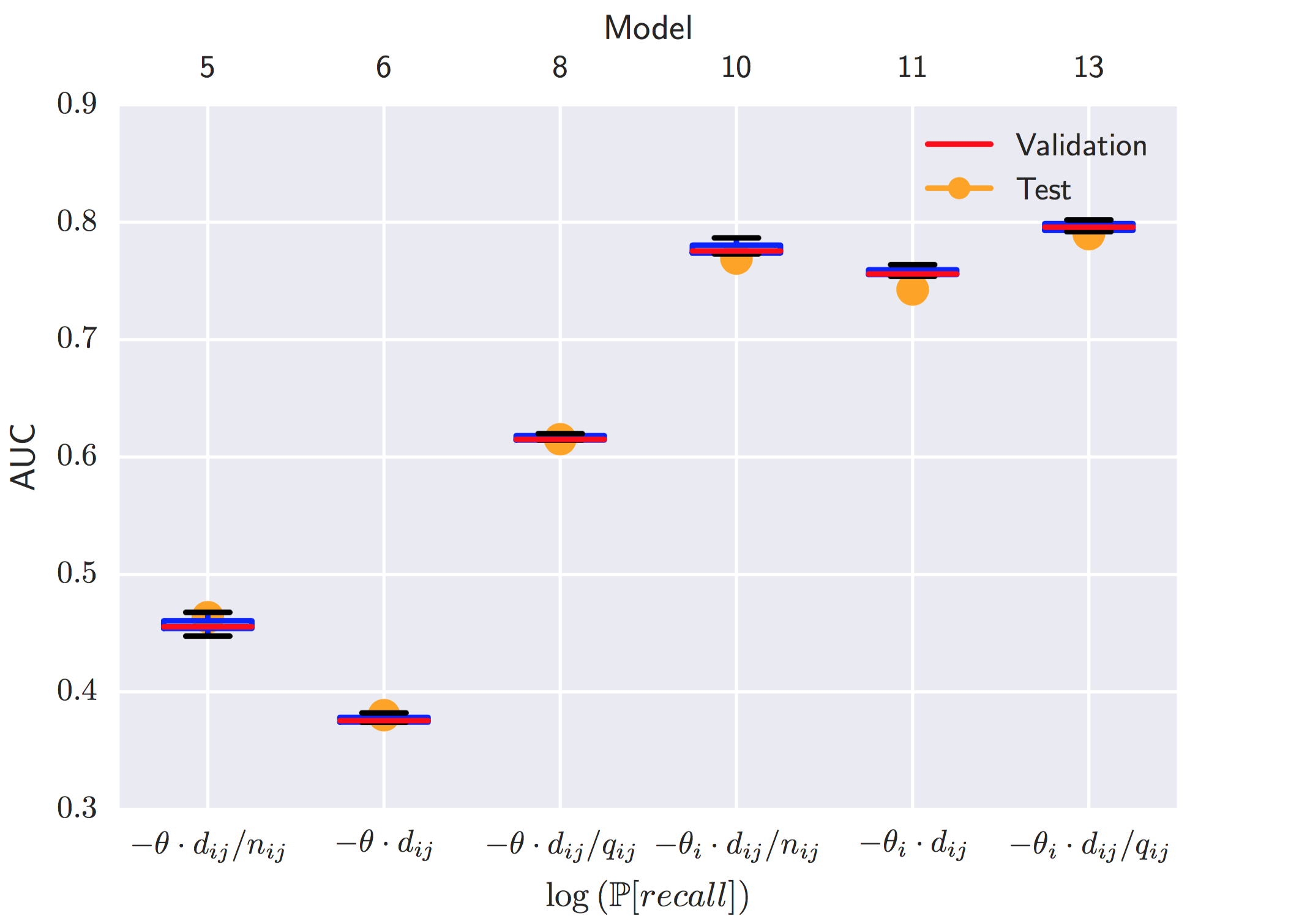}
\caption{To compare the three memory strength models $s = n_{ij}$, $s = 1$, and $s = q_{ij}$, we compute AUCs for the full data set (instead of separate bins of data, as in Fig. \ref{fig:auc-vs-deck}). The box-plots show the spread of validation AUC across the ten folds of cross-validation, and the orange circles show AUC on the test set.}
\label{fig:val-auc-rows-10-11-13}
\end{subfigure}
\end{figure}

We compare the exponential forgetting curve from Eqn. \ref{eq:expcurve} to three benchmark models: the zero- and one-parameter logistic item response theory models (henceforth, 0PL-IRT and 1PL-IRT) and logistic regression. 
The 0PL-IRT \emph{user} model assumes the recall likelihood follows $\mathbb{P}[\mbox{recall}] = \phi(\theta_j)$ for each user $j$ observed in the training set (where $\phi$ is the logistic link function); similarly, the 0PL-IRT \emph{item} model assumes $\mathbb{P}[\mbox{recall}] = \phi(\beta_i)$ for each item $i$ in the training set. 
The 1PL-IRT model \cite{linden1997handbook}, a mathematical formulation of the Rasch cognitive model \cite{rasch1993probabilistic}, has the following recall likelihood: $\mathbb{P}[\mbox{recall}] = \phi(\theta_j - \beta_i)$ for user $j$ and item $i$, where $\theta$ is user proficiency and $\beta$ is item difficulty. 
The logistic regression model uses the following statistics of the previous review 
intervals and outcomes to predict recall: mean, median, min, max, range, length, first, and last.

Logistic regression and 1PL-IRT are trained using MAP estimation with an L2 penalty, where the regularization constant is selected to maximize log-likelihood on a validation set. 
All other models are trained using maximum-likelihood estimation (i.e., with an implicit uniform prior on model parameters). 
The hyperparameters in the IRT models are user abilities $\vec{\theta}$ and/or item difficulties $\vec{\beta}$; the hyperparameters in the exponential forgetting models are item-specific difficulties $\vec{\theta}$ or a global difficulty $\theta$. 
We use ten-fold cross-validation to evaluate the memory models on the task of predicting held-out outcomes. 
Our performance metric is area under the ROC curve (AUC), which measures the discriminative ability of a binary classifier that assigns probabilities to class membership.\footnote{We use AUC as a metric instead of raw prediction accuracy because it is insensitive to class imbalance.} 
On each fold, we train on the full histories of 90\% of user-item pairs and the truncated histories of 10\% of user-item pairs, and validate on the interactions immediately following the truncated histories. 
Truncations are made uniformly at random -- see Fig. \ref{fig:pred-task} for an illustration of this scheme. 
After using cross-validation to perform model selection, we evaluate the models on a held-out test set of truncated user-item histories (20\% of the user-item pairs in the complete data set) that was not visible during the earlier model selection phase.

Table \ref{mnemosyne-results} summarizes all the models that were evaluated, with rows 1-4 representing the benchmarks and rows 5-14 variants of the exponential forgetting curve model. 
We compare models which use a global item-difficulty parameter $\theta$ (rows 5-9) vs. item-specific difficulties $\theta_i$ (rows 10-14); moreover, we allow the memory strength to be constant (rows 6 and 11), proportional to the number of reviews $n_{ij}$ (rows 5,7,10,12), or proportional to the position of the item in the Leitner system $q_{ij}$ (rows 8,9,13,14).

The predictive performance of the models on validation and test data is given in Fig. \ref{fig:auc-vs-deck} and \ref{fig:val-auc-rows-10-11-13}. We make four key observations:

\begin{enumerate}[leftmargin=*]
\item \emph{Positive impact of delay term}: Incorporating a delay term improves the performance of the memory model. In Fig. \ref{fig:auc-vs-deck}, the solid lines (with delay term) and dashed lines (without delay term) of same color encode comparable models with and without the delay term (model 5 vs.\ 7, 8 vs.\ 9, 10 vs.\ 12, 13 vs.\ 14).
\item \emph{Use of item-specific difficulties}: Item-specific difficulties $\theta_i$ outperform global item difficulty $\theta$ for lower decks ($q_{ij} \leq 2$) and higher decks ($q_{ij} > 5$), but the global difficulty performs better for intermediate decks (model 5 vs.\ 10, 8 vs.\ 13); in Fig. \ref{fig:auc-vs-deck}, compare the solid cool colors (models with $\theta$) to solid warm colors (models with $\theta_i$). 
\item \emph{Leitner position vs. number of reviews}: Setting the memory strength $s$ to be equal to the Leitner deck position $q_{ij}$ performs better than setting it to be proportional to the number of past reviews $n_{ij}$, which in turn is better than using a constant $s$; see Fig. \ref{fig:val-auc-rows-10-11-13} (model 8 vs.\ 5-6, and 13 vs.\ 10-11).
\item \emph{Performance w.r.t. benchmark models}:  Exponential forgetting models that include the delay term (models 5, 8, 10, and 13) perform comparably to 1PL-IRT (model 3), which is the best-performing benchmark model; in Fig. \ref{fig:auc-vs-deck}, compare the solid black line (model 3) to the other solid lines (models 5, 8, 10, and 13).
\end{enumerate}

Based on these observations, our model of the Leitner system uses the following exponential forgetting curve:
\begin{eqnarray}
\mathbb{P}[\mbox{recall}] = \exp{\left(-\theta \cdot d_i / q_{i} \right)}, \label{eq:memmodel}
\end{eqnarray}
where for item $i$, $d_{i}$ is the time since last reviewed, $q_i$ is its current deck in the Leitner system, and $\theta$ is the global difficulty.
The choice of the former two parameters follows from observations 3 and 4. 
The choice between using $\theta_i$ or $\theta$ is less clear from the data, and we settle on the latter due to considerations of practicality ($\theta_i$ may be unknown and/or difficult to estimate) and mathematical tractability. 
We discuss extensions of our model to using $\theta_i$ in later sections.

\section{A Stochastic Model for Spaced Repetition Systems}
\label{sec:queuemodel}

Based on the memory model developed in the previous section (as summarized in Eqn. \ref{eq:memmodel}), we now present a stochastic model for a spaced repetition system, and outline how we can use it to design good review scheduling policies.
We note again that all existing schemes for assigning review frequencies to decks in the Leitner system, and in fact, in all other spaced repetition systems, are based on heuristics with no formal optimality guarantees. 
One of our main contributions is to provide a principled method for determining appropriate schedules for spaced repetition systems.

We focus on a regime where the learner wants to memorize a very large set of items -- in particular, the number of available items is much larger than the potential amount of time spent by the learner in memorizing them. 
A canonical example of such a setting is learning words in a foreign language. 
From a technical point of view, this translates to assuming that new items are always available for introduction into the system, similar to an \emph{open queueing system} (i.e., one with a potentially infinite stream of arrivals). Moreover, this allows us to use the \emph{steady-state} performance of the queue as a measure of performance of the scheduler as an appropriate metric in this setting. 
We refer to this as the \emph{long-term learning} regime. 

As mentioned before, our model is based on the Leitner system \cite{leitner1974so}, one of the simplest and oldest spaced repetition systems.
It comprises of a series of $n$ decks of flashcards, indexed as $\{1,2,\ldots,n\}$, where new items enter the system at deck 1, and items upon being reviewed either move up a deck if recalled correctly or down if forgotten. 
In principle, the deck numbers can extend in both directions; in practice however, they are bounded both below and above -- we follow this convention and assume that items in deck $1$ are \emph{reflected} (i.e., they remain in deck $1$ if they are incorrectly reviewed), and all items which are recalled at deck $n$ (which in experiments we take as $n = 5$), are declared to be `mastered' and removed from the system.
For simplicity of exposition, we also assume that the difficulty parameter $\theta$ is the same for all items (i.e., model 8 in Table \ref{mnemosyne-results}), but we will discuss later how to allow for item-specific difficulties (i.e., model 13 in Table \ref{mnemosyne-results}).

\subsection{The Leitner Queue Network}
\label{ssec:leitnerqueue}

\begin{figure}[t]
\centering
\includegraphics[width=0.8\linewidth]{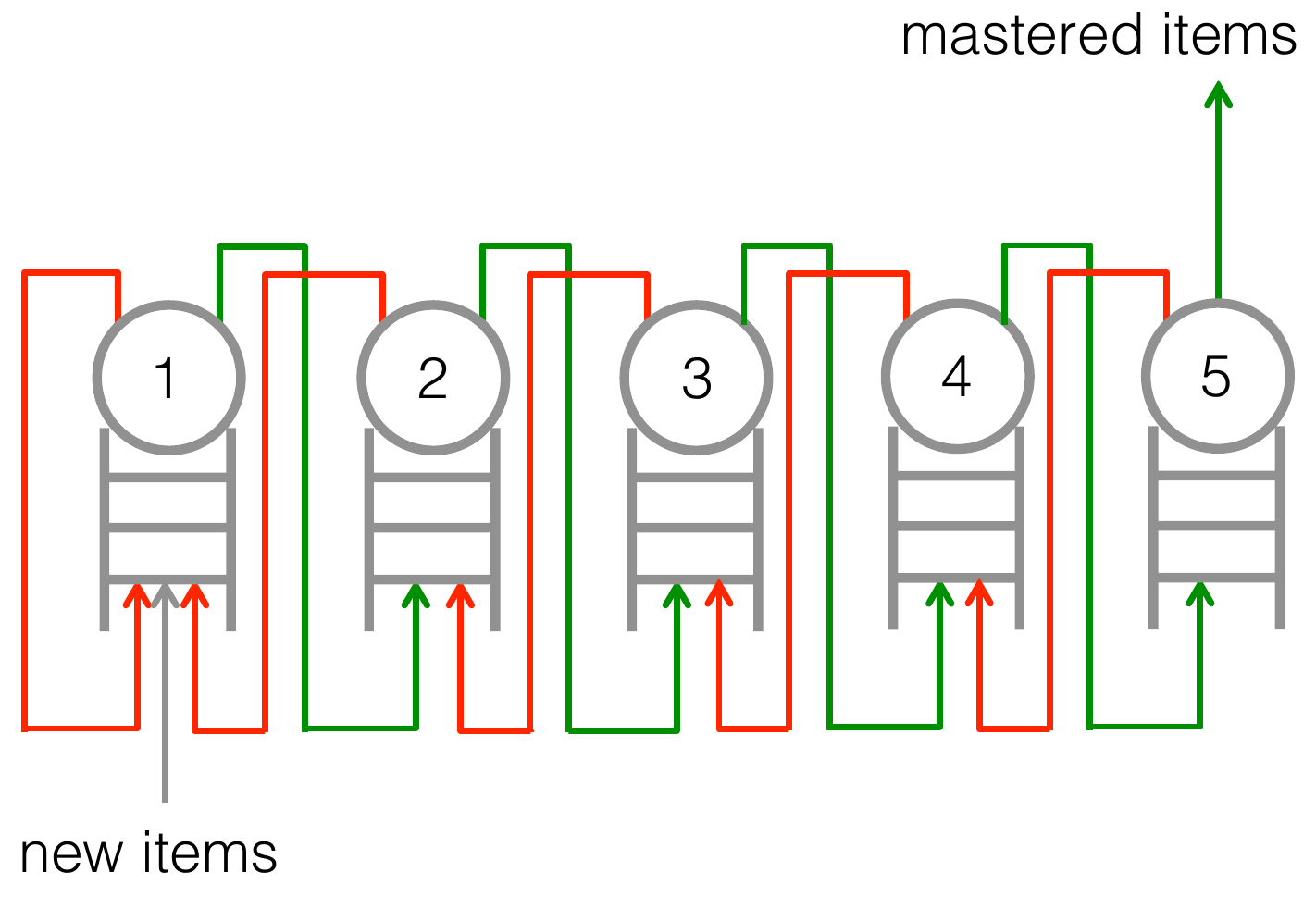}
\caption{The Leitner Queue Network: Each queue represents a deck in the Leitner system. New items enter the network at deck 1. Green arrows indicate transitions that occur when an item is correctly recalled during review, and red arrows indicate transitions for incorrectly recalled items. Queue $k$ is served (i.e., chosen for review) at a rate $\mu_k$, and selects items for review in a FIFO manner. 
}
\label{fig:lqn-diagram}
\end{figure}

We model the dynamics of flashcards in an $n$-deck Leitner system using a network of $n$ queues, as depicted in Fig. \ref{fig:lqn-diagram}. 
Formally, at time $t$, we associate with each deck $k$ the vector $S_k(t) = (Q_k(t),\{T_{k,1}(t),T_{k,2}(t),\ldots,T_{k,Q_k}(t)\})$, where $Q_k$ is the number of items in the deck at time $t$, and $T_{k,j}<t$ is the time at which the $j^{th}$ item in deck $k$ first entered the system (note that the times are sorted). 
A new item is introduced into the system at a time determined by the scheduler -- it is first shown to the user (who we assume has not seen it before), and then inserted into deck 1. 

We assume that the learner has a review frequency budget (e.g., the maximum rate at which the user can review items) of $U$, which is to be divided between reviewing the decks as well as viewing new items.
Formally, we assume that review instances are created following a Poisson process at rate $U$.
Our aim is to design a scheduler which at each review instant chooses an item to review. 
When a deck is chosen for review, we assume that items are chosen from it following a FIFO discipline. 
When an item comes up for review, its transition to the next state depends on the recall probability, which depends on the deck number and delay (i.e., time elapsed since the last review of that item). 
In particular, at time $t$, for any deck $k$, let $D_k=t-T_{k,1}$ denote the delay (i.e., the time elapsed since that item was last reviewed) of the head-of-the-line (HOL) item in deck $k$.
Then, using the memory model from Eqn. \ref{eq:memmodel}, we have that upon reviewing the HOL item from deck $k$ (for $k\in\{1,2,3,\ldots,n-1\}$), its transition follows:
\begin{align*}
\mathbb{P}[k\rightarrow k+1] &= \exp{(-\theta \cdot D_k/k)}\\
\mathbb{P}[k\rightarrow \max\{k-1,1\}] &= 1-\exp{(-\theta \cdot D_k/k)}
\end{align*}
Note that items in deck $1$ return to the same deck upon incorrect recall.
Finally, items coming up for review from deck $n$ exit the system with probability $\exp{(-\theta \cdot D_n/n)}$ (i.e., upon correct recall), else transition to deck $n-1$. 
We define the \emph{learning rate} $\lambda_{out}$ to be the long-term rate of mastered items exiting from deck $n$, i.e.: 
$$\lambda_{out} = \lim_{T\rightarrow\infty}\frac{1}{T}\cdot \big|\{\mbox{Items mastered in interval } [0,T]\}\big| $$
The aim of a scheduling policy is to maximize $\lambda_{out}$.

Given any scheduling policy that depends only on the state $\mathbf{S}(t) = (S_1(t),S_2(t),\ldots,S_n(t))$, it can be easily verified that $\mathbf{S}(t)$ forms a Markov chain.
The most general scheduler-design problem is to choose a \emph{dynamic state-dependent schedule}, wherein review instances are created following a Poisson process at rate $U$, and at each review instant, the scheduler defines a map from the state $\mathbf{S}(t)$ to a control decision which involves choosing either to introduce a new card, or a deck from which to review an item. 
Analyzing such a dynamic schedule is difficult as the state space of the system is \emph{very high dimensional} (in fact, it is infinite dimensional unless we impose some restriction on the maximum queue size). 
However, we can simplify this by restricting ourselves to \emph{static scheduling policies}: we assume that new items are injected into deck $1$ following a Poisson process with rate $\lambda_{ext}$ (henceforth referred to as the \emph{arrival rate}), and for each deck $k$, we choose a \emph{service rate} $\mu_k$, which represents the rate at which items from that deck come up for review.
We need to enforce that the arrival rate $\lambda_{ext}$ and deck service rates together satisfy the user's review frequency budget constraint, i.e., $\lambda_{ext} + \sum_k \mu_k \leq U$~\footnote{In practice, this corresponds to the following: for each review instance, with probability $\frac{\mu_k}{\lambda_{ext} + \sum_k \mu_k }$, we review the oldest item in deck $k$; else, we introduce a new item.}.

Restricting to static schedulers greatly reduces the problem dimensionality -- the aim now is to choose $\lambda_{ext},\{\mu_k\}_{k\in[n]}$ so as to maximize the learning rate $\lambda_{out}$.
We henceforth refer to this system as the \emph{Leitner Queue Network}.
The use of such static policies is a common practice in stochastic control literature, and moreover, such schedules are commonly used in designing real Leitner systems (although the review rates are chosen in a heuristic manner).
However, although the above model is potentially amenable to simulation optimization techniques, a hurdle in obtaining a more quantitative understanding is that the Markov chain over $\mathbf{S}(t)$ is \emph{time-inhomogeneous}, as the transition probabilities change with $t$.
In the next section, we propose a heuristic approximation that lets us obtain a tractable program for optimizing the review schedule.

\subsection{The Mean-Recall Approximation}
\label{ssec:approx}

The stochastic model in Section \ref{ssec:leitnerqueue} captures all the important details of the Leitner system -- however, its time-inhomogenous nature means that it does not admit a closed-form program for optimizing the review schedule.
In this section, we propose an approximation technique for the Leitner Queue Network, under which the problem of choosing an optimal review schedule \emph{reduces to a low-dimensional deterministic optimization problem}. 
Simulation results in Section \ref{ssec:analysis} (see Fig. \ref{fig:clocked-vs-expected-delay}) suggest that the two models match closely for small arrival rates. 
We note however that this approximation is essentially a heuristic, and obtaining more rigorous approximations for the proposed Leitner Queue Network model is an important topic for future work.

The main idea behind converting the model in Section \ref{ssec:leitnerqueue} to a time-homogeneous model is that for small values of $\lambda$, for which the system appears \emph{stable} (i.e., the total number of packets in the system does not blow up), then \emph{each Leitner deck $k$ behaves similarly to an $M/M/1$ queue} with service rate $\mu_k$ (chosen by the scheduler), and some appropriate input rate $\lambda_k$. 
Recall that for an $M/M/1$ queue with input rate $\lambda$ and service rate $\mu$, the total sojourn time for any packet is distributed as $Exponential(\mu - \lambda)$ (see \cite{kelly2011reversibility}).
Based on this, we assume that \emph{for an item from deck $k$ coming under review, the recall likelihood is given by}:
\begin{align}
\label{eq:meanrecall}
\mathbb{P}[\mbox{Recall}|\mbox{ Deck }k] = \mathbb{E}\left[e^{-\frac{\theta}{k} \cdot D_k}\right] = \frac{\mu_k - \lambda_k}{\mu_k - \lambda_k + \theta/k}
\end{align}
The above expression follows from the moment generating function of the exponential distribution. 
We henceforth refer to this as the \emph{mean-recall approximation}.

Formally, we define the mean-recall approximation as follows: 
Suppose we choose a static schedule $\lambda_{ext},\{\mu_k\}_{k\in[n]}$ (with $\lambda_{ext}<\mu_1$), and in addition choose input rates $\lambda_k<\mu_k$ at each deck.
Moreover, suppose the probability $P_k$ that an item from deck $k$ is recalled correctly upon review is given by Eqn. \ref{eq:meanrecall}~\footnote{One way to view this is that for each item in deck $k$ coming up for review, we ignore the true delay $D_k$ and \emph{independently generate} $\widehat{D}_k\sim Exponential(\mu_k-\lambda_k)$, which is then used to determine the recall probability $P_k$.}.
Finally, we assume the arrival rates $\{\lambda_k\}$ satisfy the following \emph{flow-balance equations}:
\begin{align*}
\lambda_1 &= \lambda_{ext} + (1-P_{1})\lambda_1 + (1-P_{2})\lambda_2\\
\lambda_{i} &= P_{i-1}\lambda_{i-1} + (1-P_{i+1})\lambda_{i+1}\quad,\mbox{for }i\in\{2,3,\ldots,n-1\}\\
\lambda_n &= P_{n-1}\lambda_{n-1},
\end{align*}

Under the above assumptions, the Leitner Queue Network is a \emph{Jackson network} of $M/M/1$ queues $Q_k$ ~\cite{kelly2011reversibility, kendall1953stochastic}, with arrival rate $\lambda_k$ and service times $\{\mu_k\}$, and from Jackson's theorem, we have that all queues are ergodic, and in steady-state, for each deck $k$, the sojourn time $D_k$ is indeed distributed as $Exponential(\mu_k-\lambda_k)$. Moreover, ergodicity also gives that the learning rate $\lambda_{out}$ is the same as the external injection rate $\lambda_{ext}$.
Putting everything together, we get the following \emph{static planning problem}:
\begin{eqnarray}
\underset{\{\mu_k\}_{k=1}^n}{\text{Maximize}} & & \lambda_{ext} \label{eq:statplan}\\
\text{Subject to} \!\!& & U\geq \lambda_{ext} + \sum\limits_{k=1}^n \mu_k , \nonumber \\
&& \lambda_1 =  \lambda_{ext} + (1 - P_{1})\lambda_1 + (1-P_{2})\lambda_2, \nonumber\\
&& \,\lambda_{k} = P_{k-1}\lambda_{k-1} + (1-P_{k+1})\lambda_{k+1}\;,\mbox{for }k\!\neq\! 1,\!n, \nonumber\\
&&\lambda_n = P_{n-1}\lambda_{n-1}, \nonumber\\
&&P_{k} = \frac{\mu_k - \lambda_k}{\mu_k - \lambda_k + \theta/k}\quad\,\forall k\in[n], \nonumber\\
&& \,0\leq \lambda_k\leq\mu_k\quad\,\forall k\in[n], \nonumber
\end{eqnarray}
Thus, as desired, the mean-recall approximation helps convert the stochastic control problem of designing an optimal review schedule to a low ($O(n)$) dimensional, deterministic optimization problem.
Now, we can use a nonlinear solver (e.g., IP-OPT) to solve the static planning problem. 
Note that our problem is unusual compared to typical network control problems, as the \emph{routing probabilities depend on the service rates $\mu_k$}. 
Also, note that ergodicity of the system is critically dependent on the conditions $~\lambda_k < \mu_k\quad,\forall k\in\{1,2,\ldots,n\}$ -- if, however, one or more of these do not hold, then the resulting queue length(s) (i.e., deck sizes), and thus, the delays between reviews for items in that deck, grow unbounded for the decks for which the condition is violated. 
Moreover, since items move to lower decks upon being incorrectly recalled, choosing a high injection rate should result in items building up in the lowest deck. 
We verify these qualitative observations through experiments in Section \ref{sec:experiments}.

\subsection{Features of Optimal Leitner Schedules}
\label{ssec:analysis}

\begin{figure}[t]
\centering
\includegraphics[width=0.95\linewidth]{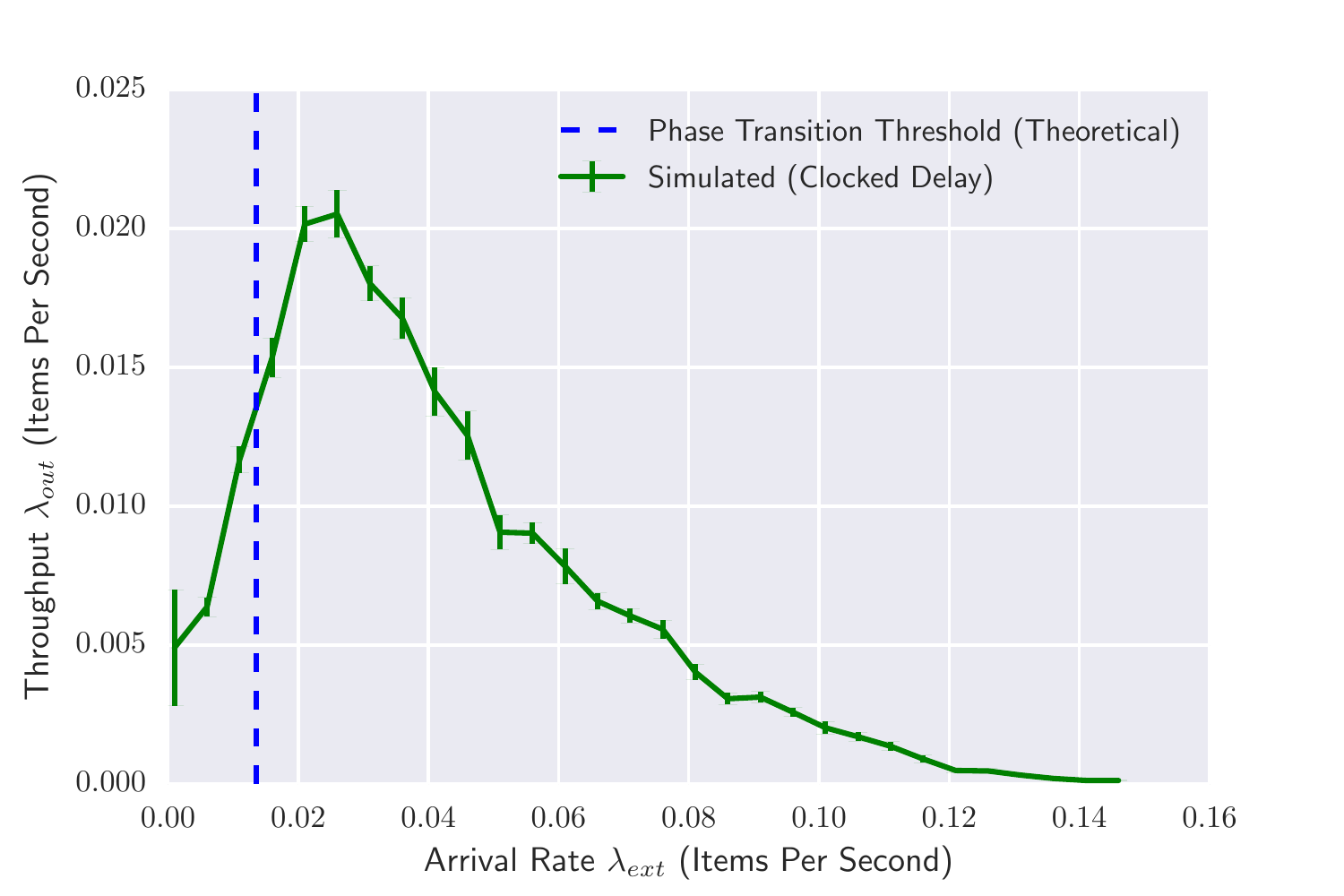}
\caption{Phase transition in learning: Average learning rate $\lambda_{out}$ vs. arrival rate $\lambda_{ext}$ in the Leitner Queue Network with clocked delays, for a session of $500$ reviews over $50$ items. We set number of decks $n = 5$, review frequency budget $U = 0.1902$, and global item difficulty $\theta = 0.0077$. The dashed vertical line is the predicted phase transition threshold under the mean-recall approximation.}
\label{fig:phase-transition}
\end{figure}

We now explore the properties of the optimal review schedule for the Leitner Queue Network under the mean-recall approximation. 
The main qualitative prediction from our model is the \emph{existence of a phase transition} in learning outcomes: Given a schedule $\{\mu_k\}$, there is a threshold $\lambda_t$ s.t. for all $\lambda_{ext}>\lambda_t$, there are no feasible solutions $\{\lambda_k\}$ satisfying Eqn. \ref{eq:statplan}. Moreover, if $\lambda_{ext}>\lambda_t$, then the lowest Leitner deck (i.e., $Q_1$) experiences packet accumulation and delay blow-up, and thus the learning rate $\lambda_{out}$ goes to $0$. 
In Fig. \ref{fig:phase-transition}, we simulate a review session with 500 reviews and 50 unique items for different values of $\lambda_{ext}$. 
We observe that a sharp phase transition indeed occurs as the arrival rate is increased: throughput initially increases linearly with arrival rate, then sharply decreases. 

\begin{figure}[t]
\centering
\includegraphics[width=0.95\linewidth]{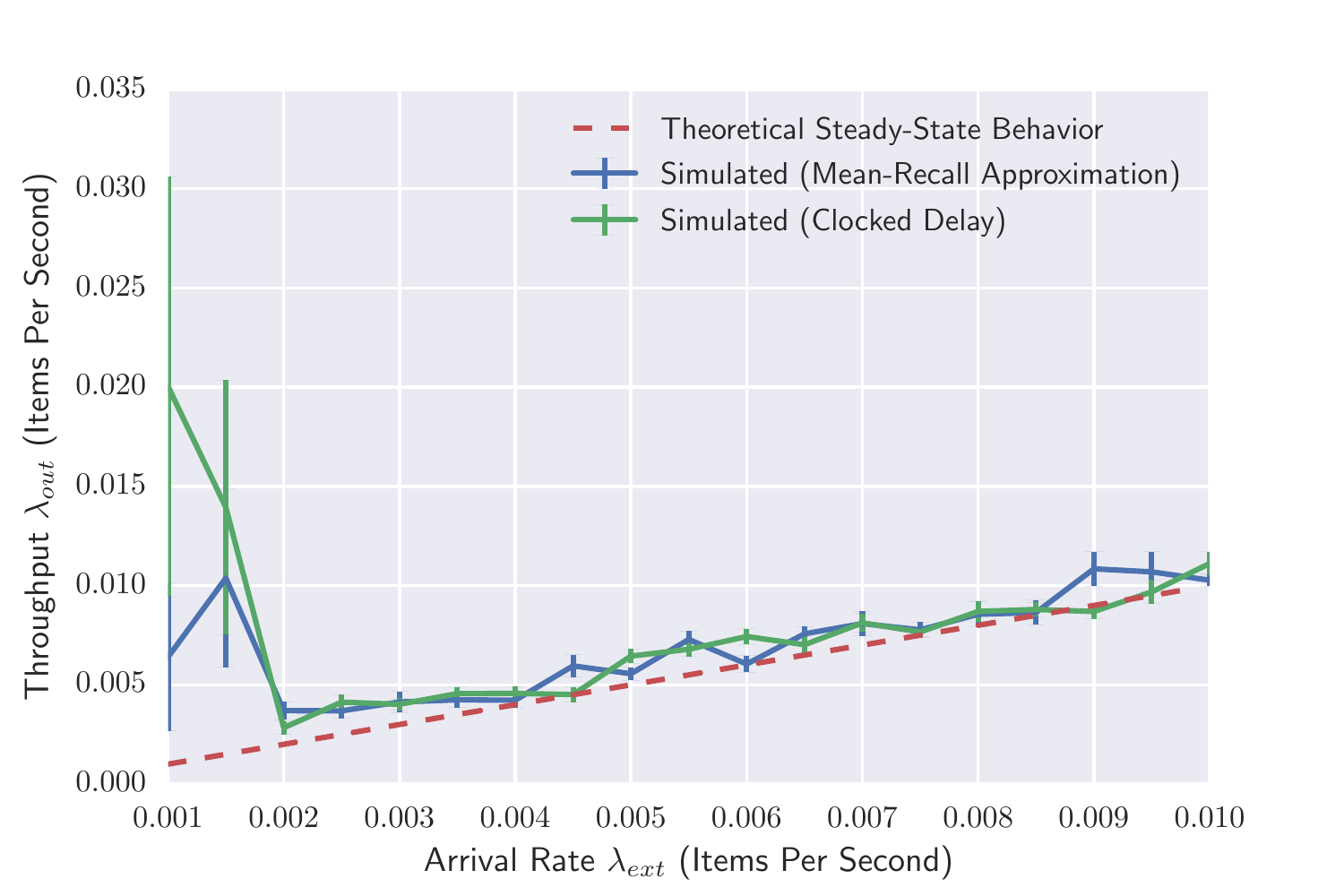}
\caption{The mean-recall approximation: Average learning rate $\lambda_{out}$ vs. arrival rate $\lambda_{ext}$, for $500$ reviews over $50$ items. We set number of decks $n = 5$, review frequency budget $U = 0.1902$, and global item difficulty $\theta = 0.0077$. The green curve is generated using clocked delays, while the blue curve uses the mean-recall approximation. The $\lambda_{out} = \lambda_{ext}$ line (red dashed curve) is the steady-state $\lambda_{out}$ under the mean-recall approximation.}
\label{fig:clocked-vs-expected-delay}
\end{figure}

The simulation in Fig. \ref{fig:phase-transition} is for the Leitner Queue Network with actual (or \emph{clocked}) delays, i.e., item routing is based on actual times between reviews.
The dotted line indicates the phase transition threshold obtained under our mean-recall approximation (Eqn. \ref{eq:statplan}), which appears to be a lower bound (i.e., a conservative estimate) for the true phase transition point for review sessions of moderate length. 
Fig. \ref{fig:clocked-vs-expected-delay} verifies our intuition that the mean-recall approximation performs well for small values of $\lambda_{ext}$. 
Obtaining more rigorous guarantees on the approximation remains an open question.

\begin{figure}
\begin{subfigure}
\centering
\includegraphics[width=\linewidth]{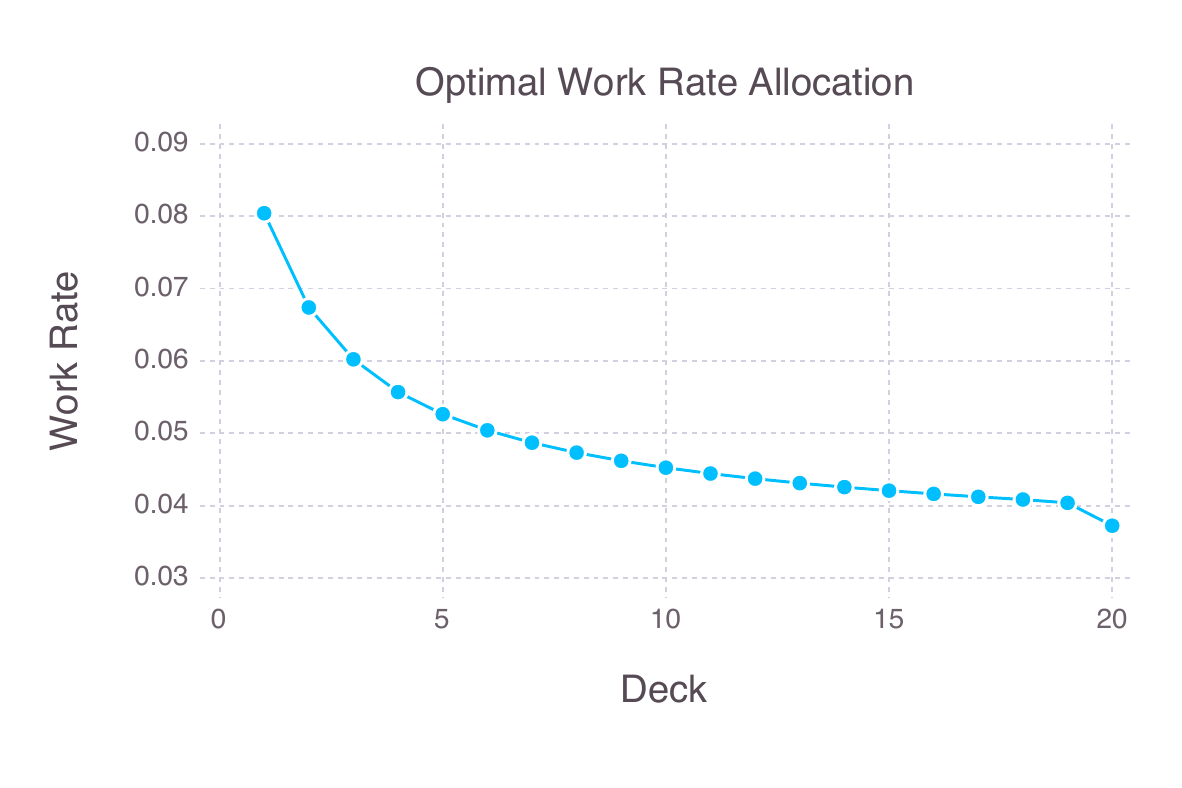}
\caption{Optimal review schedule $\{\mu_k\}_k$ for $n = 20$, $U = 1$, $\theta = 0.01$}
\label{fig:p1}
\end{subfigure}
\begin{subfigure}
\centering
\includegraphics[width=\linewidth]{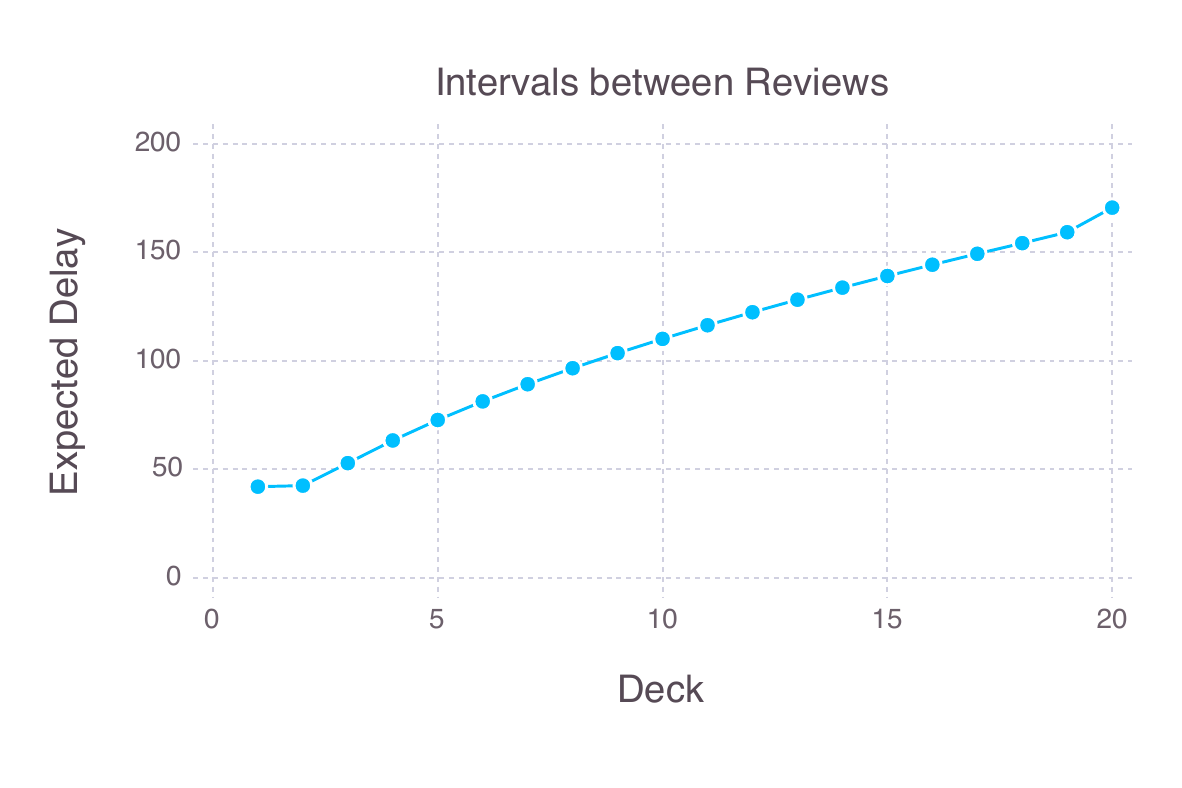}
\caption{Expected delays $1/(\mu_k - \lambda_k)$ under optimal schedule for $n = 20$, $U = 1$, $\theta = 0.01$.}
\label{fig:p2}
\end{subfigure}
\begin{subfigure}
\centering
\includegraphics[width=\linewidth]{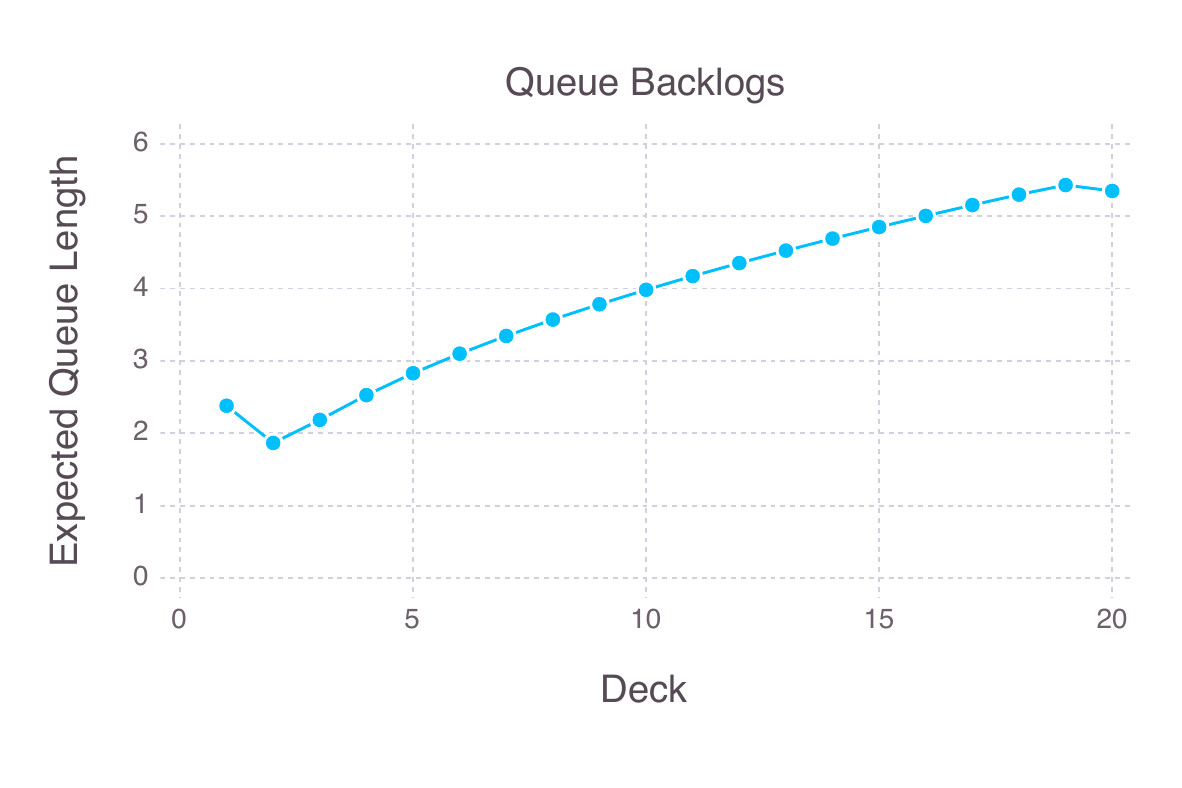}
\caption{Expected queue size $\lambda_k/(\mu_k - \lambda_k)$ under optimal schedule for $n = 20$, $U = 1$, $\theta = 0.01$. The kinks at the boundaries, in this and the previous plot, arise from having a bounded number of decks.}
\label{fig:p3}
\end{subfigure}
\end{figure}

The above simulations suggest that the mean-recall approximation gives a good heuristic for optimizing the learning rate.
Moreover, the tractability of the resulting  optimization program (Eqn. \ref{eq:statplan}) lets us investigate structural aspects of the optimal schedule under the mean-recall assumption.
In Fig. \ref{fig:p1}, we see that the optimal schedule spends more time on lower decks than on higher decks (i.e., $\mu_k\leq \mu_{k+1}\,\forall\,k$). 
This is partly a result of the network topology, where items enter the system through deck 1 and exit the system through deck $n$. 
However, in Fig. \ref{fig:p2} we observe that the Leitner Queue Network also increases the expected delay between subsequent reviews as an item moves up through the system.
Note that longer review intervals does not follow from decreasing $\mu_k$, as the (steady-state) deck sizes can be different, as is indeed the case (see Fig. \ref{fig:p3}).
We note here that there is empirical support in the literature for expanding intervals between repetitions \cite{cepeda2006distributed}. 

\begin{figure}
\begin{subfigure}
\centering
\includegraphics[width=\linewidth]{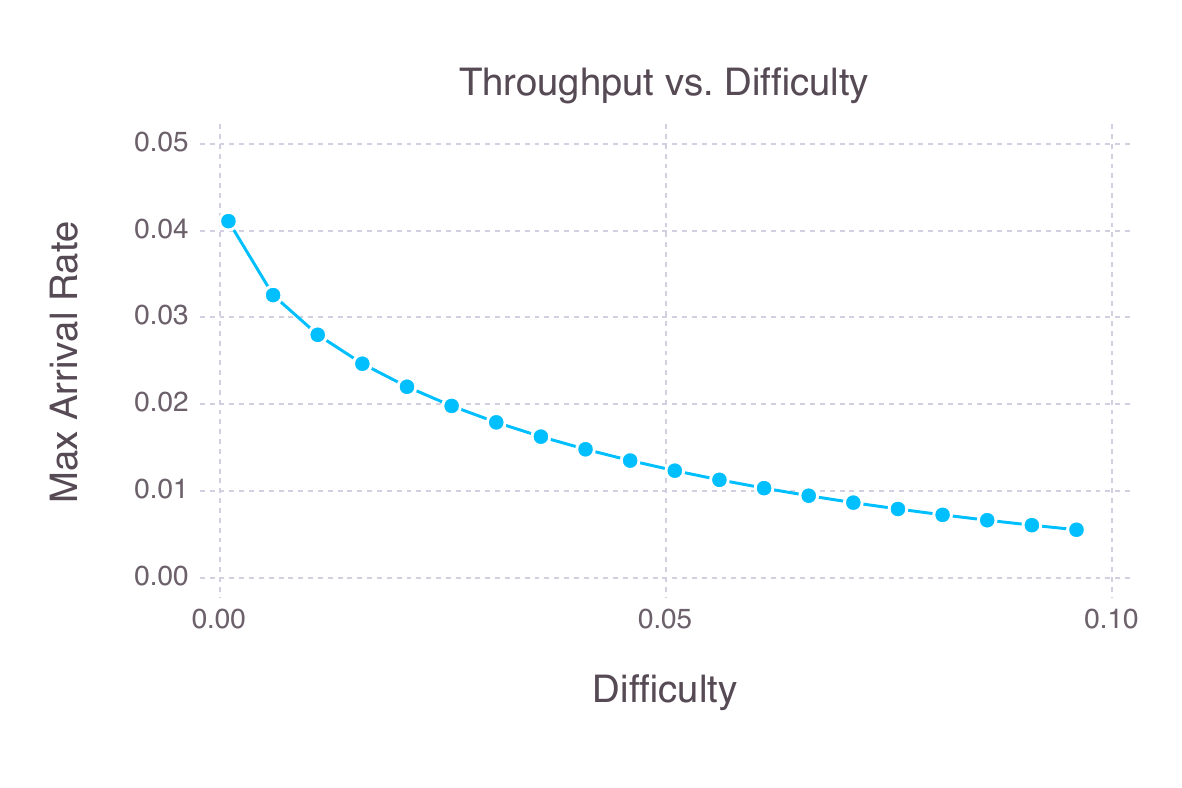}
\caption{Variation in maximum learning rate $\lambda_{ext}^\ast$ with item difficulty $\theta$, for $n = 20, U = 1$.}
\label{fig:p5}
\end{subfigure}
\begin{subfigure}
\centering
\includegraphics[width=\linewidth]{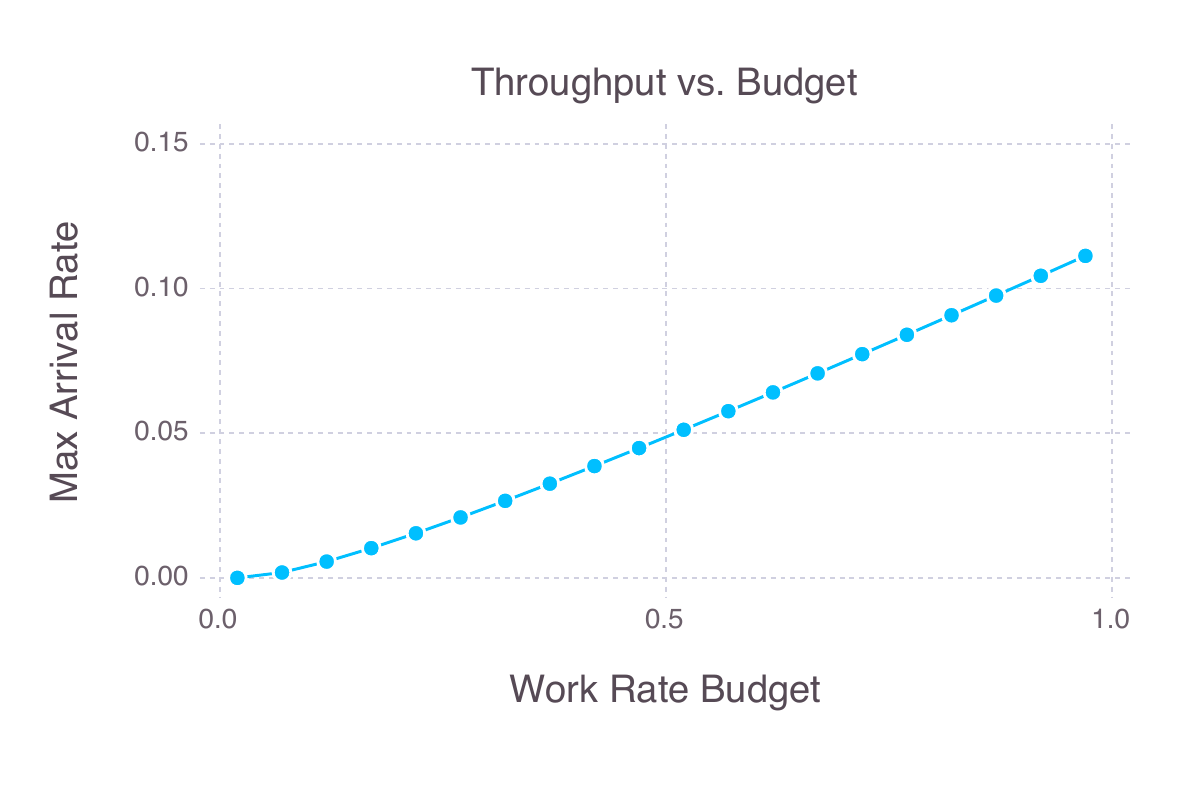}
\caption{Variation in learning rate $\lambda_{ext}^\ast$ with review frequency budget $U$ for $n = 5, \theta = 0.01$. Note the convexity (hence, increasing returns) for low $U$.}
\label{fig:p6}
\end{subfigure}
\begin{subfigure}
\centering
\includegraphics[width=\linewidth]{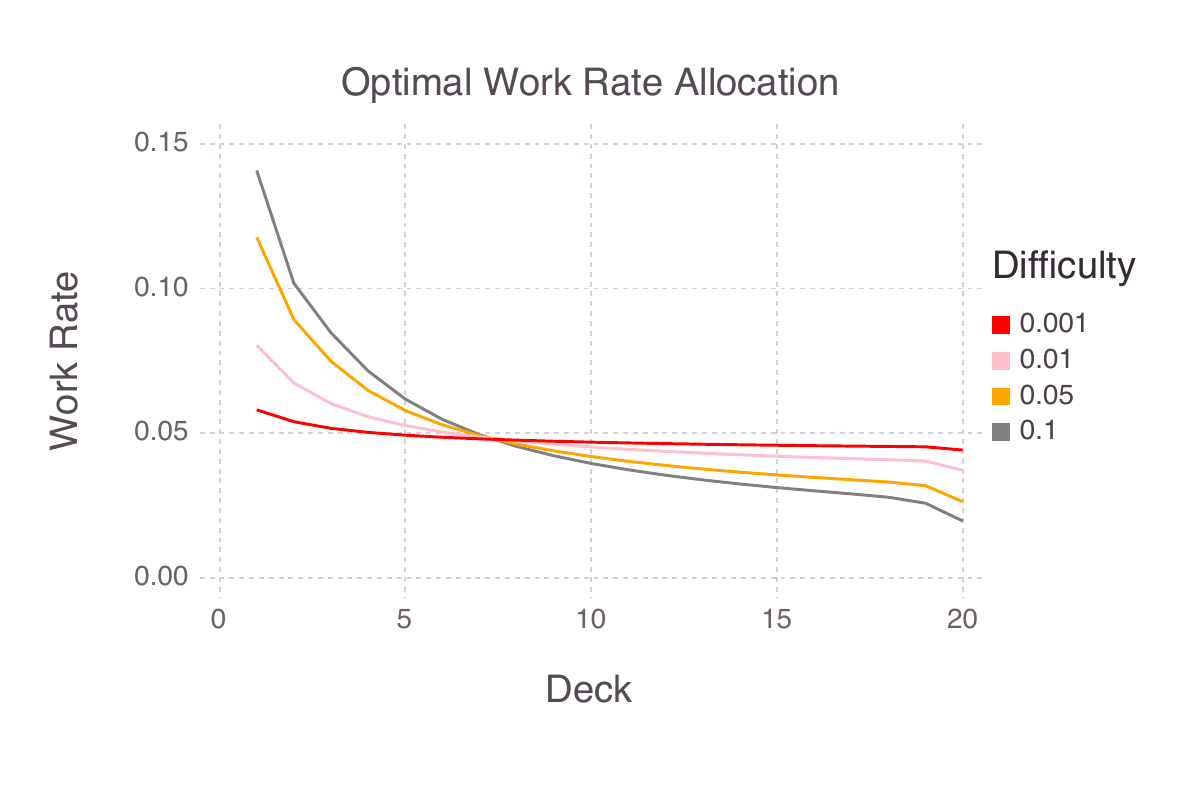}
\caption{Optimal review schedule $\{\mu_k\}_k$ for $n = 20$, $U = 1$ and varying item difficulties $\theta$.}
\label{fig:p4}
\end{subfigure}
\end{figure}

Finally, Fig. \ref{fig:p5} and \ref{fig:p6} show how the maximum achievable learning rate depends on the general difficulty of items, and the user's review frequency budget $U$. The convexity of the latter plot is encouraging, as it suggests that there are \emph{increasing} returns (for lower $U$) as the user increases their budget. 


\subsection{Extension to item-specific difficulties}

The assumption that all items have the same difficulty $\theta$ can be relaxed by discretizing difficulties into a fixed number of bins $b$, and creating $b$ parallel copies of the network. 
We now have $b$ parallel Leitner Queue Networks, coupled via the budget constraint which applies to the sum of service rates across the networks for each $\theta$.
We can assume that the $\theta_i$ are known a priori (e.g., from log data or expert annotations).
To understand the effect of different $\theta_i$, we compare the optimal schedule for different $\theta_i$ (but using the same budget $U$) in Fig. \ref{fig:p4}. 
The result is interesting, because to the best of our knowledge, there is little understanding of how the user should adjust deck review frequencies when the general difficulty of items changes. 
Fig. \ref{fig:p4} suggests that when items are generally easy, the user should spend a roughly uniform amount of time on each deck; however, when items are of higher general difficulty, the user should spend more time on lower decks than higher decks.

\section{Experimental Validation}
\label{sec:experiments}

To empirically test the fidelity of the Leitner Queue Network as a model for spaced repetition, we perform an experiment on Amazon Mechanical Turk (MTurk) involving participants memorizing words from a foreign language. Our study is designed to experimentally verify the existence of the phase transition (shown in Fig. \ref{fig:phase-transition}), the primary qualitative prediction made by our model.

\subsection{Experiment Setup}

\begin{figure}[t]
\centering
\includegraphics[width=0.8\linewidth]{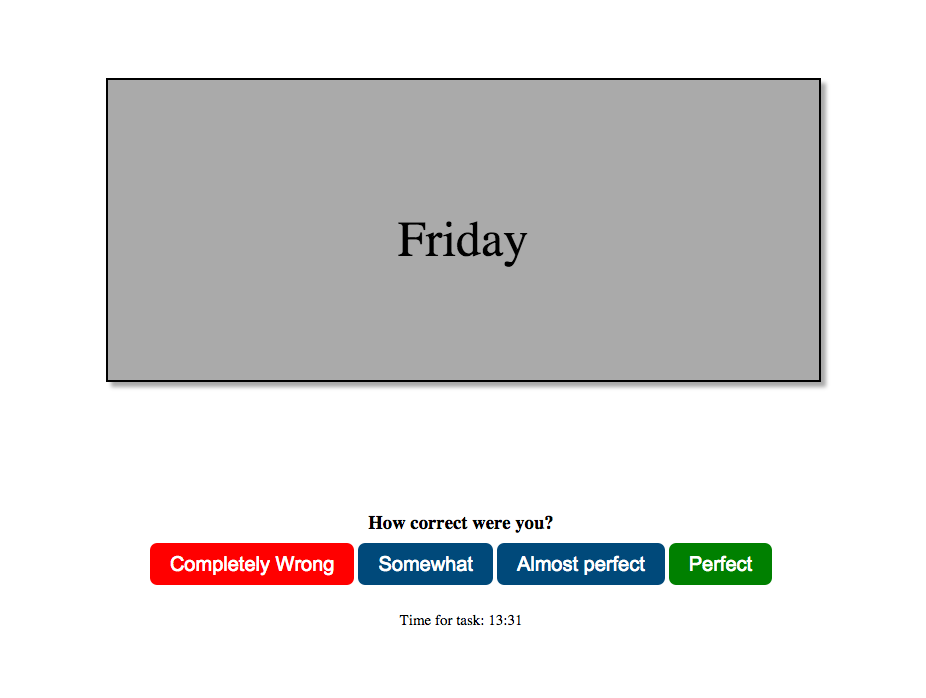}
\caption{Screenshot of our Mechanical Turk interface : The user sees a word in \emph{Japanese} (not shown) and enters a guess. The card is then flipped to reveal the word's meaning in English (shown above), and the user then assesses herself.}
\label{fig:mturk-screen}
\end{figure}

\begin{figure}
\begin{subfigure}
\centering
\includegraphics[width=\linewidth]{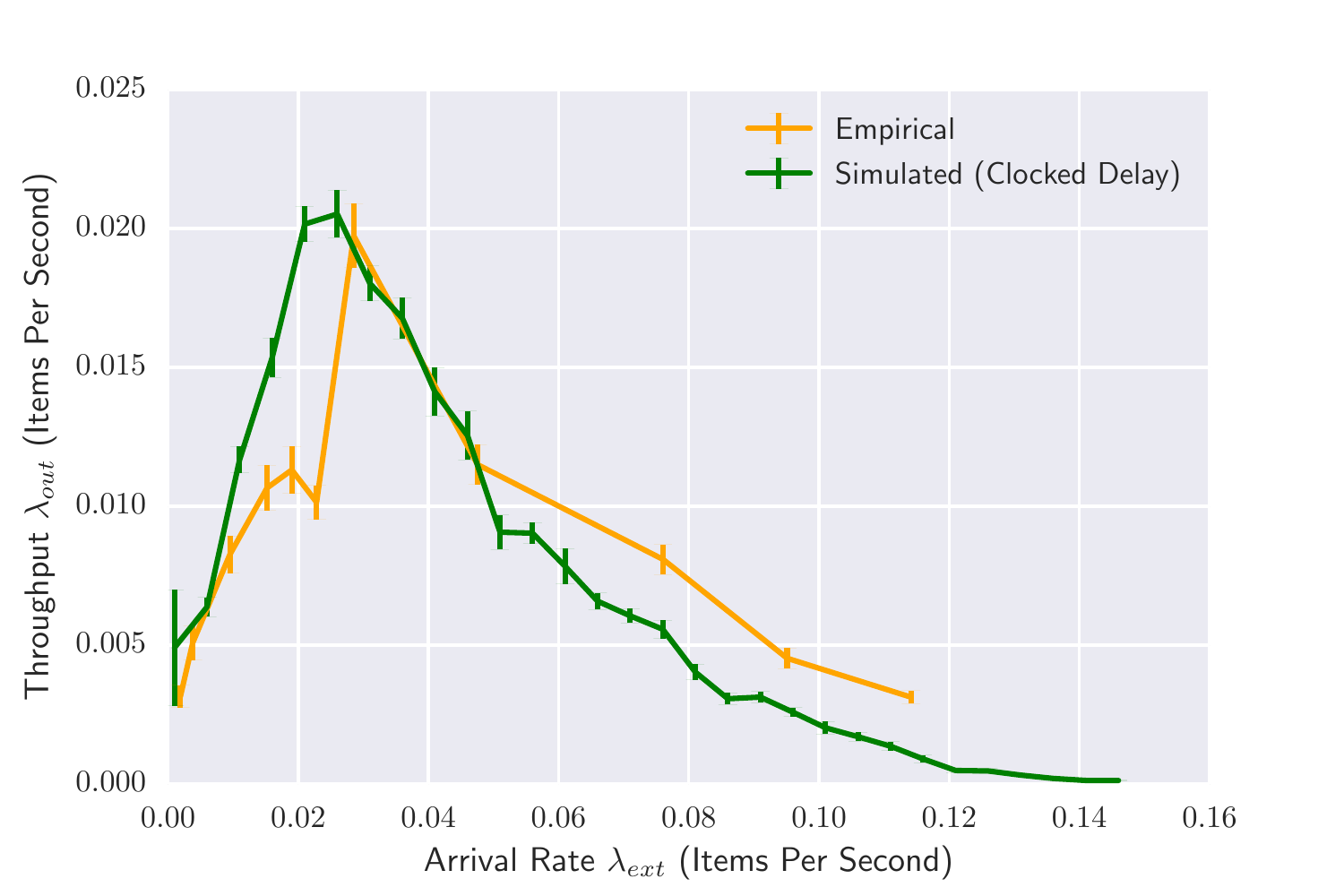}
\caption{The exit rate $\lambda_{out}$ vs. arrival rate $\lambda_{ext}$, where number of decks $n = 5$, review frequency budget $U = 0.1902$, and global item difficulty $\theta = 0.0077$.}
\label{fig:phase-transition-post-mturk}
\end{subfigure}
\begin{subfigure}
\centering
\includegraphics[width=\linewidth]{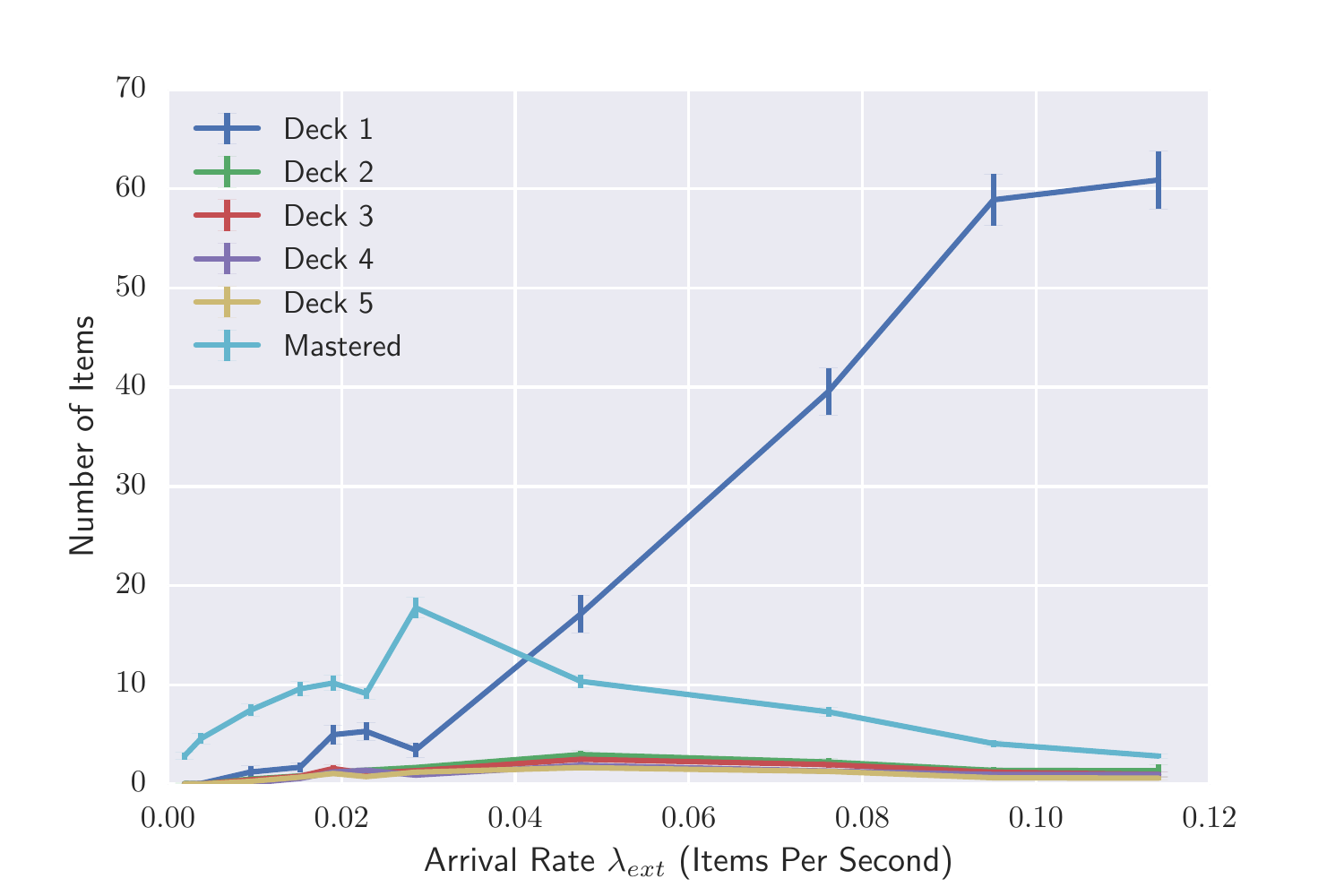}
\caption{The number of items that finish in each deck vs. arrival rate $\lambda_{ext}$, where number of decks $n = 5$, review frequency budget $U = 0.1902$, and global item difficulty $\theta = 0.0077$.}
\label{fig:phase-transition-deck-distrns}
\end{subfigure}
\begin{subfigure}
\centering
\includegraphics[width=\linewidth]{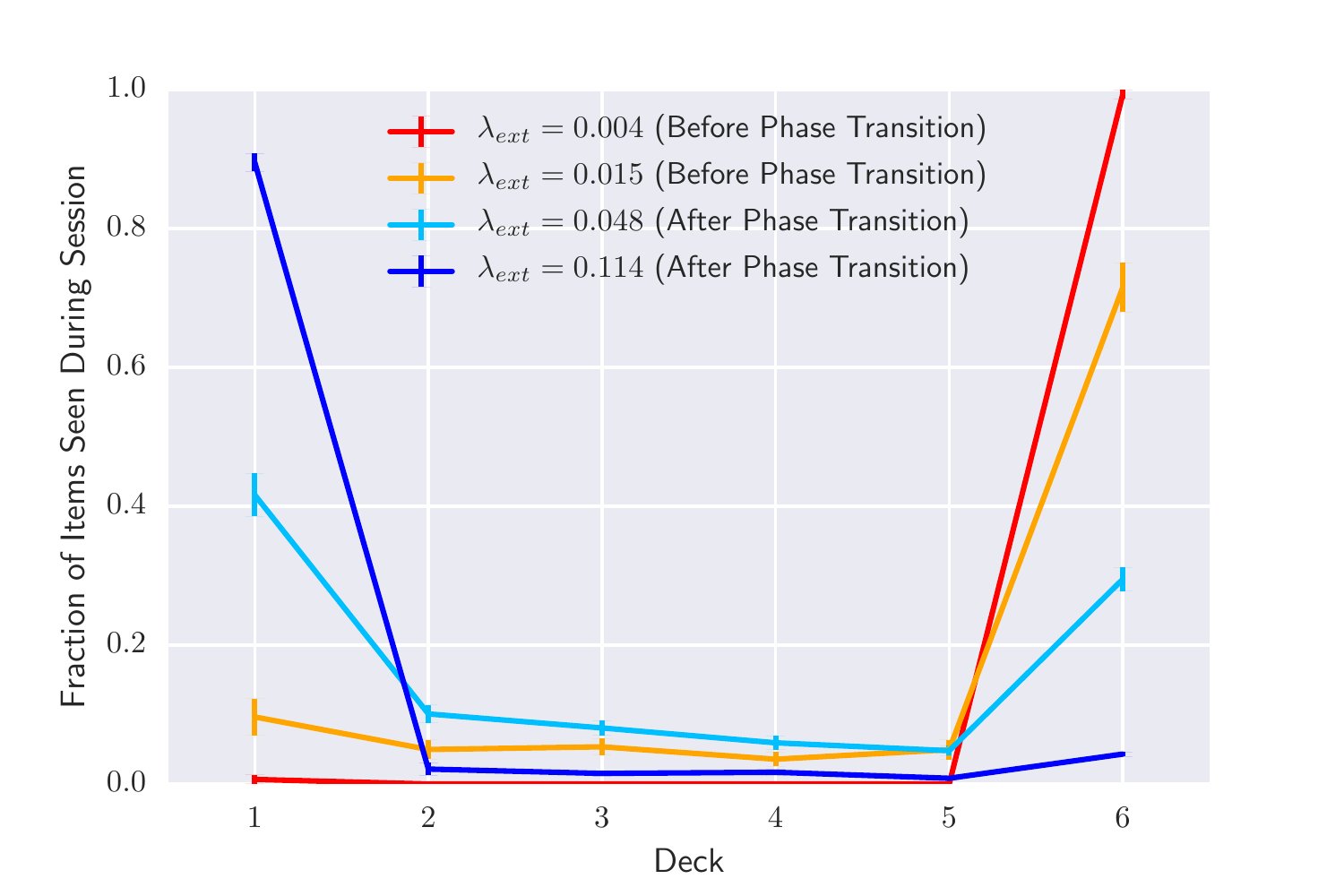}
\caption{The fraction of items seen during a session that finish in each deck for different arrival rates $\lambda_{ext}$, where number of decks $n = 5$, review frequency budget $U = 0.1902$, and global item difficulty $\theta = 0.0077$. Deck 6 refers to the pile of mastered items.} 
\label{fig:phase-transition-deck-distrns-lambda-lines}
\end{subfigure}
\end{figure}

A total of 331 users (`turkers') on the MTurk platform were solicited to participate in a vocabulary learning task. At the beginning of the task, vocabulary used in the task was selected randomly from one of two categories: \emph{Japanese} (words) and \emph{American Sign Language} (animated gestures). Items for the experiment were sampled from the list of common words in both languages \footnote{\url{https://en.wiktionary.org/wiki/Appendix:1000_Japanese_basic_words} for \emph{Japanese} and \url{http://www.lifeprint.com/asl101/gifs-animated/} for \emph{American Sign Language}}. Each task was timed to last 15 minutes, and turkers were compensated \$1.00 for completing the task regardless of their performance.  The task used the interface depicted in Fig. \ref{fig:mturk-screen}: a flashcard initially displaying the item in the foreign language (either word or gesture), and a pair of \emph{YES/NO} buttons to collect input from the user in response to the question, \emph{``Do you know this word?''}. If the user selected \emph{YES}, she was then asked to type the translation of the word in English. Once the word was entered, the flashcard was ``flipped'', revealing the correct English word. The user was then asked to self-assess their correctness on a scale of 1 (completely wrong) to 4 (perfect). Following the submission of the rating, the next card was sampled from the deck. In all our experiments, we consider a self-assessment score of 3 (almost perfect) or 4 (perfect) as a ``pass'', and all other scores as a ``fail''. 

At the beginning of each task, each turker was assigned to one of the 11 conditions corresponding to the arrival rate ($\lambda_{ext}$) of new items: [0.002,  0.004,  0.010,  0.015,  0.020, 0.023,  0.029,  0.050,  0.076,  0.095, 0.11,  0.19] (items per second). The resulting data set consists of a total of $77,034$ logs, $331$ unique users, $446$ unique items, an overall recall rate of $0.663$, a fixed session duration of 15 minutes, and an average session length of $171$ logs, where each log is a tuple (turkerID, itemID, score, timestamp). We set deck review rates to $\mu_k \propto 1 / \sqrt{k}$ -- this roughly follows the shape of the optimal allocation in Fig. \ref{fig:p1}. We note that choosing an optimal scheduler is not essential for our purpose of observing the phase transition -- moreover, we cannot optimize the review rates $\mu_k$ ex ante as we do not know the item difficulty $\theta$ or the review budget $U$. During the experiments, we set the number of decks in the system to $n = \infty$, so items never exit the system during a review session. Items incorrectly recalled at deck 1 are `reflected' and stay in deck 1. In our post-hoc analysis of the data, we consider an item to be `mastered' if its final position is in deck $6$ or greater. 

We estimate the empirical review budget $U$ as (average number of logs in a session) / (session duration), and the empirical item difficulty $\theta$ using maximum-likelihood estimation. We measure throughput $\lambda_{out}$ as (average number of items mastered in a session) / (session duration). 

\subsection{Results}

Fig. \ref{fig:phase-transition-post-mturk} overlays the empirical learning rate from the MTurk data for each arrival rate condition on the learning rate curve for the simulated Leitner Queue Network (same as Fig. \ref{fig:phase-transition}) using the parameter values for $\theta$ and $U$ measured from the MTurk data (see caption of Fig. \ref{fig:phase-transition} for details). 
The simulated and empirical curves are in close agreement; in particular, the MTurk data shows the phase transition in learning rate predicted by our theoretical model.

In addition to computing the observed throughput for the various arrival rates in the MTurk data, we compute the average distribution of items across the five decks and the pile of mastered items at the end of a session. This gives insight into where items accumulate in unstable regimes. 
Fig. \ref{fig:phase-transition-deck-distrns} illustrates the same phase transition observed earlier: as the arrival rate increases, we first see an increase in the number of mastered items. 
However, as the arrival rate increases past the optimum, relatively fewer items are mastered and relatively more items get `stuck' in deck 1. Intuitively, the user gets overwhelmed by incoming items so that fewer and fewer items get reviewed often enough to achieve mastery. 
Fig. \ref{fig:phase-transition-deck-distrns} and \ref{fig:phase-transition-deck-distrns-lambda-lines} match the behavior suggested by our queueing model: for injection rates higher than the threshold, the number of items in deck 1 blows up while the other decks remain stable.

\section{Conclusion and Open Questions}
\label{sec:conclusions}

Our work develops the first formal mathematical model for reasoning about spaced repetition systems that is validated by empirical data and provides a principled, computationally tractable algorithm for flashcard review scheduling. Our formalization of the Leitner system suggests the maximum speed of learning as a natural design metric for spaced repetition software; using techniques from queueing theory, we derive a tractable program for calibrating the Leitner system to optimize the speed of learning. Our queueing framework opens doors to leveraging an extensive body of work in this area to develop more sophisticated extensions. To inspire and facilitate future work in this direction, we release (1) all model and evaluation code, (2) framework code for carrying out user studies, and (3) the data collected in our Mechanical Turk study. The data and code for replicating our experiments are available online at \url{http://siddharth.io/leitnerq}.

Our work suggests several directions for further research. The primary follow-up is to obtain a better understanding of the Leitner Queue Network; in particular, better approximations with rigorous performance guarantees.
Doing so will allow us to design better control policies, which ideally could maximize the learning rate in the transient regime.
The latter is critical for designing policies for \emph{cramming}~\cite{novikoff2012education}, a complementary problem to long-term learning where the number of items to be learnt is of the same order as the number of reviews. 
Next, our queueing model should be modified to incorporate more sophisticated memory models that more accurately predict the effect of a particular review schedule on retention.
Finally, there is a need for more extensive experimentation to understand how closely these models of spaced repetition apply to real-world settings.

\section{Acknowledgements}

We thank Peter Bienstman and the Mnemosyne project for making their data set publicly available. This research was funded in part through NSF Awards IIS-1247637, IIS-1217686, IIS-1513692, the Cornell Presidential Research Scholars Program, and a grant from the John Templeton Foundation provided through the Metaknowledge Network. 

\bibliographystyle{abbrv}
\bibliography{master}

\end{document}